\documentclass[format=acmsmall, review=false, screen=true]{acmart}

\usepackage{booktabs} 
\usepackage{subfig}
\usepackage[ruled]{algorithm2e} 
\usepackage{soul}
\newcommand{\etal}{\textit{et al.}}

\graphicspath{{./figures/}}

\SetAlFnt{\small}
\SetAlCapFnt{\small}
\SetAlCapNameFnt{\small}
\SetAlCapHSkip{0pt}
\IncMargin{-\parindent}

\acmJournal{TIST}
\acmVolume{9}
\acmNumber{4}
\acmArticle{39}
\acmYear{2010}
\acmMonth{3}
\copyrightyear{2009}

\setcopyright{acmlicensed}

\acmDOI{0000001.0000001}

\received{February 2007}
\received[revised]{March 2009}
\received[accepted]{June 2009}

\begin{document}
\title[s]{Online Tensor-Based Learning   for Multi-Way  Data}

\author{Ali Anaissi}
\affiliation{%
  \institution{ School of Computer Science, The University of Sydney}
  \city{Camperdown}
  \state{NSW}
  \postcode{2006}
  \country{Australia}}
\email{ali.anaissi@sydney.edu.au}

\author{Basem Suleiman}
\affiliation{%
	\institution{ School of Computer Science, The University of Sydney}
	\city{Camperdown}
	\state{NSW}
	\postcode{2006}
	\country{Australia}}
\email{basem.suleiman@sydney.edu.au}

\author{Seid Miad Zandavi}
\affiliation{%
	\institution{ School of Computer Science, The University of Sydney}
	\city{Camperdown}
	\state{NSW}
	\postcode{2006}
	\country{Australia}}
\email{miad.zandavi@sydney.edu.au}

\begin{abstract}	
		The online analysis of multi-way data stored in  a tensor  $\mathcal{X}  \in  \mathbb{R} ^{I_1 \times \dots   \times I_N} $   has become an essential tool for capturing the  underlying structures and extracting the sensitive features which can be used to learn a predictive model. However, data distributions often evolve with time and a current predictive model may not be sufficiently representative in the future. Therefore,  incrementally updating  the tensor-based features and model coefficients are required  in such situations.  A new efficient tensor-based feature extraction, named NeSGD, is proposed for online  $CANDECOMP/PARAFAC$ (CP) decomposition. According to the new features obtained from the resultant matrices of NeSGD, a new criteria is triggered for the updated process of the online predictive model. Experimental evaluation in the field of structural health monitoring using laboratory-based and real-life structural  datasets show that our methods provide more accurate results compared with existing online tensor analysis and model learning. The results showed that the proposed methods significantly improved the classification error rates, were able to assimilate the changes in the positive data distribution over time, and maintained a high predictive accuracy in all case studies.

\end{abstract}

%
%


\maketitle

\renewcommand{\shortauthors}{A. Anaissi et al.}

\section{Introduction}

Almost all major cities around the world have developed complex physical infrastructure that encompasses bridges, towers, and iconic buildings. One of the most emerging challenges with such infrastructure is to continuously monitor its health to ensure the highest levels of safety. Most of the existing structural monitoring and maintenance approaches rely on a time-based visual inspection and manual instrumentation methods which are neither efficient nor effective. 

Internet of Things (IoT) has created a new paradigm for connecting things (e.g., computing devices, sensors and objects) and enabling data collection over the Internet. Nowadays, various types of IoT devices are widely used in smart cities to continuously collect data that can be used to manage resources efficiently. For example, many sensors are connected to bridges to collect various types of data about their health status. This data can be then used to monitor the health of the bridges and decide when maintenance should be carried out in case of potential damage  \cite{Li2015SHMBridges}. With this advancement, the concept of smart infrastructure maintenance has emerged as a continuous automated process known as Structural Health Monitoring (SHM) \cite{Jinping2010SHMReview}. 

SHM provides an economic monitoring approach as the inspection process which is mainly based on a low-cost IoT data collection system. It also improves effectiveness due to the automation and continuity of the monitoring process. SHM enhances understanding the behaviour of structures as it continuously provides large data that can be utilized to gain insight into the health of a structure and make timely and economic decisions about its maintenance. 

One of the critical challenges in SHM is the non-stationary nature of the data collected from several networked sensors attached to a bridge \cite{Xin2018Nonstationary}. This collected data is the foundation for training a model to detect potential damages in a bridge. In SHM, the data used in training the model comes from healthy samples (it does not include damaged data set) and hence represents one-class data. Also, the data is collected within a fixed  period time and processed simultaneously. This influences the performance of model training as it does not consider data variations over more extended period time. Structures always experience severe events such as heavy loads over a long and continuous period of time and high and low seasonal temperatures. Such variations would dramatically affect the status of the data as it changes the characteristics of the structure such as the data frequencies. Consequently, the training data fed into the model does not represent such variations but only healthy or undamaged data samples. Thus it is critical to develop a method that mitigates these variations and  increases the specificity rate.

\begin{figure*}
	\centering
	\includegraphics[scale=0.33]{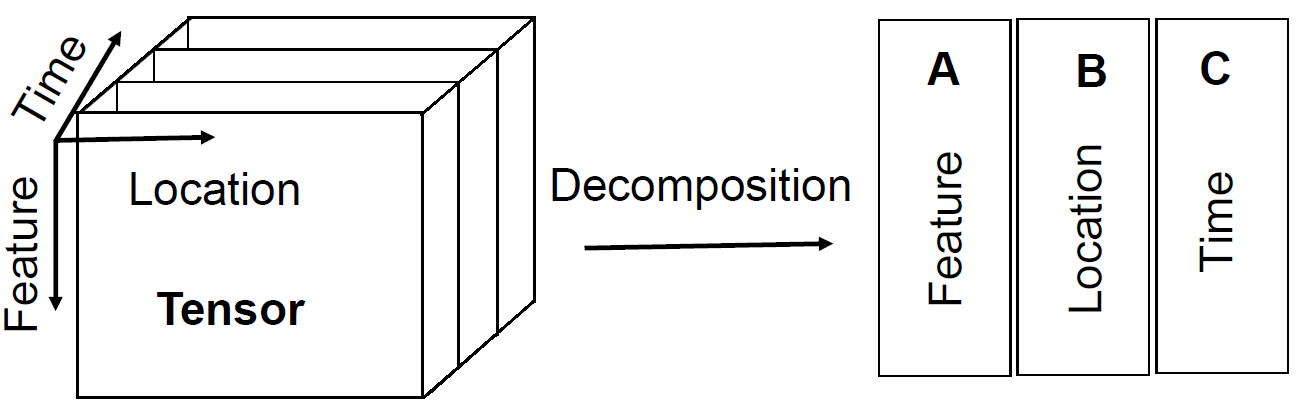}
	\caption{Tensor data with three modes in SHM}
	\label{tensor}
\end{figure*}

Another challenge in SHM is that generated sensing data exists in a correlated multi-way form which makes the standard two-way matrix analysis unable to capture all of these correlations and relationships \cite{Cheema2016twoway}. Instead, the SHM data can be arranged into a three-way data structure where each data point is triple of a feature value extracted from a particular sensor at a specific time. Here, the information extracted from the raw signals in time domain represent features, the sensors represent data location and time snapshots represent the timestamps when data was extracted (as shown in Figure \ref{tensor}).

In order to address the problem as mentioned above,  an online one-class learning approach should be employed along with an online  multi-way data analysis tool to capture the underlying structure inherits in multi-way sensing data. In this setting, the design of a  one-class support vector machine (OCSVM) \cite{anaissi2017adaptive} and tensor analysis  are well-suited to this kind of problems where only observations from the multi-way positive (healthy) samples are required.

Tensor  is a multi-way extension of the  matrix to represent multidimensional  data structures such that SHM data. Tensor analysis requires extensions to the  traditional two-way data analysis methods such as Non-negative
Matrix Factorization (NMF), Principal Component Analysis (PCA) and Singular Value Decomposition (SVD). In this sense, the \textbf{CANDECOMP/PARAFAC} (CP) \cite{Phan2013Candecomp} has  recently become a standard approach for analyzing multi-way tensor data. Several algorithms have been proposed to solve CP  decomposition \cite{symeonidis2008tag} \cite{lebedev2014speeding} \cite{rendle2009learning}. Among these algorithms, alternating least square(ALS)  has been heavily employed which repeatedly solves each component matrix  by locking all other components until it   converges \cite{papalexakis2017tensors}. It is usually applied in offline mode to decompose the positive training tensor data, which then fed into an OCSVM model to construct the decision boundary. However, this offline process is also not suitable for such dynamically changing SHM data. Therefore, we also interested here to incrementally update the  resultant CP decomposition in an online manner.

Similarly, the OCSVM  has  become a standard approach in solving anomaly detection problems. OCSVM is usually trained with a set of positive (healthy) training data, which are collected within a fixed time period and are processed together at once. As we mentioned before, this fixed batch model generally performs poorly if the distribution of the training data varies over a time span. One simple approach would be to retrain the OCSVM model from scratch when additional positive data arrive. However, this would lead to ever-increasing training sets, and would eventually become impractical. Moreover, it also seems computationally wasteful to retrain the model for each incoming datum, which will likely have a minimal effect on the previous decision boundary. Another approach for dealing with large non-stationary data would be to develop an online-OCSVM model that incrementally updates the decision boundary, i.e., incorporating additional healthy data when they are available without retraining the model from scratch. The question now is how to distinguish real damage from  the environmental changes which require model updates. Current research (such as \cite{wang2013online,davy2006online}) proposes a threshold value to measure the closeness of a new negative datum to the decision boundary for online OCSVM. More specifically, if this new negative datum is not far from the decision boundary, then they consider it as a healthy sample (environmental changes) and update the model accordingly. However, this predefined threshold is very sensitive to the distribution of the training data and it may include or exclude anomalies and healthy samples. Recently, Anaissi \etal \cite{anaissi2017self} propose another approach which measures the similarity between a new event and error support vectors  to generate a self-advised decision value rather than using a fixed threshold. That was an effective solution but it is susceptible to include damage samples if the model keeps updated in the same direction as the real damage samples. Then the resultant updated model will start encounter damage samples in the training data. Thus this approach will start missing  real damage events.

To address the aforementioned problems, we propose a new method that uses the online learning technique to solve the problems of online OCSVM and CP decomposition. We employ stochastic gradient descent (SGD) algorithm for online CP decompositio, and we introduce a new criterion to trigger the update process of the online-OCSVM, This criterion utilities  the information derived from the location component matrix which we obtain when we decompose the three-way tensor data $\mathcal{X}$. This matrix  stores meaningful information about the behavior for each sensor location on the bridge. Intuitively, environmental changes such as temperature will affect all the instrumented sensors  on the  bridge similarly. However, real damage would affect a particular sensor location and the ones close by. The contributions  of our proposed method are as follows:

 \begin{itemize}

	\item \textbf{Online CP decomposition.}  We employ  Nesterov's Accelerated Gradient (NAG) method into SGD algorithm  to solve the CP decomposition which has the capability to update $\mathcal{X}^{(t+1)}$ in one single step. We also followed the perturbation approach which adds a little noise to the gradient update step to reinforce the next update step to start moving away from a saddle point toward the correct direction. 	
		
	\item \textbf{ Online anomaly detection.}  We propose  a tensor-based  online-OCSVM which is able to distinguish between  environmental   and damage behaviours to adequately update the model coefficients.

	\item \textbf{Empirical analysis on structural  datasets.} We conduct experimental analysis using laboratory-based and real-life  datasets in the  field of SHM. The experimental analysis shows that our method can achieve lower false alarm rates  compared to other known existing online and offline methods.

\end{itemize}

This paper is organized as follows. Section ~\ref{sec:RelatedWork} presents preliminary work and discusses research related to this work. In section ~\ref{sec:Method} we introduce the details of our method; online OCSVM-OCPD. Section ~\ref{sec:Experiement} presents the experimental evaluation of the proposed method and discuses the results. Conclusions and future work are discussed in Section ~\ref{sec:Conclusion}.

\section{ Preliminaries and Related Work}
\label{sec:RelatedWork}

Our research work builds upon and extends essential methods and algorithms including Tensor (CP) Decomposition, Stochastic Gradient Descent, and Online One-Class Support Vector Machine. We first discuss the key elements of these methods and algorithms and then follow that with an analysis of related studies and their contribution to addressing the challenges discussed in the introduction. We conclude this with the discussion with the weaknesses of existing work and how our proposed work attempts to address these weaknesses. 

\subsection{CP Decomposition}
Given a three-way tensor $\mathcal{X} \in \Re^{I \times J \times K} $,  CP  decomposes $\mathcal{X}$ into three  matrices  $A \in \Re^{I \times R}$, $B \in \Re^{J \times R} $and $ C \in \Re^{K \times R}$, where $R$ is the latent factors. It can be written as follows:
\begin{eqnarray}\label{eq:decomp}
\mathcal{X}_{(ijk)}  \approx \sum_{r=1}^{R}A_{ir} \circ B_{jr} \circ C_{kr}
\end{eqnarray}
where  "$\circ$" is a vector outer product.  $R$ is the latent element, $A_{ir}, B_{jr} $ and $C_{kr}$ are r-th columns of component 
matrices $A \in \Re^{I \times R}$, $B \in \Re^{J \times R} $and $ C \in \Re^{K \times R}$.
The main goal of  CP decomposition is   to decrease the sum  square error  between  the model and a given tensor $\mathcal{X}$. Equation \ref{eq:als} shows our loss function $L$ needs to be optimized.
\begin{eqnarray}\label{eq:als}	
L (\mathcal{X}, A, B, C) = \min_{A,B,C} \|  \mathcal{X} - \sum_{r=1}^R  \ A_{ir} \circ B_{jr} \circ C_{kr} \|^2_f,
\end{eqnarray}
where $\|\mathcal{X}\|^2_f$  is the sum squares of $\mathcal{X}$ and the subscript $f$ is the Frobenius norm. The loss function $L$ presented in Equation \ref{eq:als} is a non-convex problem with many local minima since it aims to optimize the sum squares of three matrices. The CP decomposition often uses the Alternating Least Squares (ALS) method to find the solution for a given tensor. The ALS method follows the offline training process which iteratively solves each component matrix by fixing all the other components, then it repeats the procedure until it converges \cite{khoa2017smart}. The rational idea of the least square algorithm is to set the partial derivative of the loss function to zero concerning the parameter we need to minimize. Algorithm \ref{ALS} presents the detailed steps of ALS.

\begin{algorithm}
	\caption{  Alternating Least Squares for CP}
	\label{ALS}
	\textbf{Input}: Tensor $\mathcal{X} \in \Re^{I \times J \times K}  $, number of components $R$\\
	\textbf{Output}: Matrices  $A \in \Re^{I \times R}$, $B \in \Re^{J \times R} $ and  $ C \in \Re^{K \times R}$
	\begin{enumerate}
		\item[1:] Initialize $A,B,C$
		\item[2:] Repeat
		{\setlength\itemindent{6pt}
			\item[3:] $A = \underset{A}{\arg\min} \frac{1}{2} \| X_{(1)} - A ( C \odot B)^T\|^2 $
			\item[4:] $B = \underset{B}{\arg\min} \frac{1}{2} \| X_{(2)} - B ( C \odot A)^T\|^2 $
			\item[5:] $C = \underset{C}{\arg\min} \frac{1}{2} \| X_{(3)} - C ( B \odot A)^T\|^2 $
			\item[]($X_{(i)} $ is the  unfolded matrix of $X$ in a current mode)    
		}
		\item[6:] until convergence
	\end{enumerate}
	
\end{algorithm}

In online settings, it is a naive approach would be to recompute the CP decomposition from scratch for each new incoming  $X^{(t+1)}$. Therefore, this would become impractical and computationally expensive as new incoming datum would have a minimal effect on the current tensor. Zhou et al. \cite{zhou2016accelerating} proposed a method called onlineCP to address the problem of online CP decomposition using the ALS algorithm. The method was able to incrementally update the temporal mode in multi-way data but failed for non-temporal modes \cite{khoa2017smart}. In recent years, several studies have been proposed to solve the CP decomposition using the stochastic gradient descent (SGD) algorithm which will be discussed in the following section.

\subsection{Stochastic Gradient Descent}
Stochastic gradient descent algorithm is a key tool for optimization problems. Assume that our aim is to optimize a loss function  $L(x,w)$, where $x$ is a data point  drawn from a  distribution $\mathcal{D}$ and $w$ is a variable. The stochastic optimization problem can be defined as follows:
\begin{eqnarray}\label{eq:sgd}
w =  \underset{w}{argmin} \;    \mathbb{E}[L(x,w)]
\end{eqnarray}
The stochastic gradient descent method solves the above problem defined in Equation \ref{eq:sgd} by  repeatedly updates $w$ to minimize $L(x,w)$. It starts with some initial value of $w^{(t)}$ and then repeatedly performs the update as follows:
\begin{eqnarray}\label{eq:sgdu}
w^{(t+1)} :=    w^{(t)}   + \eta   \frac{\partial L}{\partial w } (x^{(t)} ,w^{(t)} )
\end{eqnarray}
where $\eta$ is the learning rate and $x^{(t)}$ is a random sample drawn from the given distribution $\mathcal{D}$.

This method guarantees the convergence of the loss function $L$ to the global minimum when it is convex. However, it can be susceptible to many local minima and saddle points when the loss function exists in a non-convex setting. Thus it becomes an NP-hard problem. The main bottleneck here is due to the existence of many saddle points and not to the local minima \cite{ge2015escaping}. This is because the rational idea of gradient algorithm depends only on the gradient information which may have $\frac{\partial L}{\partial u } = 0$ even though it is not at a minimum. 

Recently, SGD has attracted several researchers working on tensor decomposition.   For instance,  Ge \etal   \cite{ge2015escaping} proposed a perturbed SGD (PSGD) algorithm for orthogonal tensor optimization. They presented several theoretical analysis that ensures convergence; however, the method does not apply to non-orthogonal tensor. They also didn't address the problem of slow convergence. Similarly, Maehara \etal  \cite{maehara2016expected} propose a new algorithm for CP decomposition based on a combination of SGD and ALS methods (SALS). The authors claimed the algorithm works well in terms of accuracy. Yet its theoretical properties have not been completely proven and the saddle point problem was not addressed. Rendle and Thieme \cite{rendle2010pairwise} propose a pairwise interaction tensor factorization method based on  Bayesian personalized rank. The algorithm was designed to work only on three-way tensor data. To the best of our knowledge, this is the first work that applies SGD algorithm augmented with Nesterov's optimal gradient and perturbation methods for optimal CP decomposition of multi-way tensor data.

\subsection{Online One-Class Support Vector Machine}
\label{sec:online-OCSVM}
Given a set of training data $X=\{{x_i}\}_{i=1}^n$, with $n$ being the number of samples, OCSVM maps these samples into a high dimensional feature space using a function $\phi$ through  the kernel $K(x_i,x_j) = \phi(x_i)^T \phi(x_j)$. Then OCSVM learns a decision boundary that maximally separates the training samples from the origin \cite{scholkopf2001estimating}. The primary objective of OCSVM   is to optimize  the  following equation:
\begin{eqnarray}\label{ocsvm}    
\max_{w,\xi,\rho} -\frac{1}{2} \lVert w \rVert^{2} - \frac{1}{\nu n} \sum_{i=1}^n \xi_i + \rho
\end{eqnarray}
\begin{eqnarray*}    
	s.t \hspace{2em}  w. \phi(x_i) \geq \rho - \xi_i,\hspace{1em}   \xi_i \geq 0, \hspace{1em}  i = 1, \ldots, n.
\end{eqnarray*}
where $\nu$ $ (0<\nu<1)$ is a user defined parameter  to control the rate of anomalies in the training data, $\xi_i$ are the slack variable, $\phi(x_i)$ is the kernel matrix and $ w. \phi(x_i) - \rho$ is the separating hyperplane in the feature space. The problem  turns  into a dual objective by introducing Lagrange multipliers $\alpha = \{\alpha_1,\cdots,\alpha_n\}$. This dual optimization problem is solved using the following  quadratic programming formula \cite{scholkopf2002learning}:
\begin{eqnarray}\label{quad}
W=\min_{W(\alpha,\rho)}    \frac{1}{2}\sum_{i}^n \sum_{j}^n\alpha_i\alpha_j \phi(x_i, x_j) + \rho(1 - \sum_{i}^n\alpha_i)
\end{eqnarray}
\begin{eqnarray*}
	s.t \hspace{2em} 0 \leq \alpha_i \leq 1,\hspace{1em} \sum_{i=1}^n    \alpha_i =\frac{1}{\nu n}.
\end{eqnarray*}
where $\phi(x_i, x_j)$ is the kernel matrix, $\alpha$ are the Lagrange multipliers and $\rho$ known as  the bias term.

The partial derivative of the quadratic optimization problem (defined in Equation \ref{quad}) with respect to $\alpha_i$,  $\forall i \in S$,  is then used as a decision function to calculate  the  score for  a new incoming sample:
\begin{eqnarray}\label{dv2}
g(x_i) = \frac{\partial w}{\partial \alpha_i} =  \sum_{j}\alpha_i \phi(x_i, x_{j})-\rho.
\end{eqnarray}
The OCSVM uses Equation \ref{df}  to identify whether a new incoming point belongs to the positive class when returning a positive value, and vice versa if it generates a negative value.
\begin{eqnarray}\label{df}
f(x_i) = sgn(g(x_i))
\end{eqnarray}
The achieved OCSVM solution must always satisfy the constraints from the Karush-Khun-Tucker (KKT) conditions, which are described in Equation~\ref{kkt}.
\begin{eqnarray}\label{kkt}
g(x) = \Bigg\{\begin{tabular}{ccc}
$\geq$ 0 &   & $\alpha_i=0$ \\
= 0 &  & $ 0 < \alpha_i < 1$\\
$<$ 0  &   &$   \alpha_i  = 1 $
\end{tabular}
\end{eqnarray}
where  $\alpha_i = 0$  is referred to the  non-support or reserve training vectors denoted by \textit{R}, $\alpha_i=1$  represents non-margin support or error vectors denoted by \textit{E} and  $ 0 < \alpha_i < 1$  represents the support vectors denoted by \textit{S}. 

In an online setting, we need to ensure the KKT conditions are maintained while learning and adding a new data point to the previous solution. Several researchers address the problem of online SVM \cite{cauwenberghs2000incremental,diehl2003svm,laskov2006incremental} where all are based on the original method known as bookkeeping which proposed by Cauwenberghs \etal.  The method computes the new coefficients of the SVM model while preserving the KKT conditions. This method made useful contributions to the incremental learning of two classes SVM (TCSVM), but it cannot be used for the one-class problem. Davy \textit{et al.} \cite{davy2006online} concluded that such incremental SVM methods cannot be directly applied to the one-class problem as they are dependent on the TCSVM margin areas which do not exist in OCSVM. Therefore, Davy \etal proposed an online OCSVM-based threshold value. In this method, the anomaly score is computed for each incoming datum and evaluated against a predefined threshold value. The new datum is added to the training data and the model coefficients are updated accordingly when its value is greater than the threshold. Similarly, Wang \etal \cite{wang2013online} presented an online OCSVM algorithm for detecting abnormal events in real-time video surveillance. Their algorithm combines online least-squares OCSVM (LS-OCSVM) and sparse online LS-OCSVM. The basic model is initially constructed through a learning training set with a limited number of samples. Like \cite{davy2006online} algorithm, the model is then updated through each incoming datum using threshold-based evaluation.  Recently, Anaissi \etal ~\cite{anaissi2017self} propose another approach which measures the similarity between a new event and error support vectors to generate a self-advised decision value rather than using a fixed threshold. That was an effective solution but it is susceptible to include damage samples if the model keeps updated in the same direction as the real damage samples. Then the resultant updated model will start encountering damage samples in the training data. Thus this approach will start missing the real damage events.

\section{Online Tensor-Based Learning   for Multi-Way  Data}
\label{sec:Method}

The incremental learning  of online-OCSVM has been well-studied and proved to produce the same solution as the batch learning process (offline learning). In fact, the main problem of online-OCSVM is not related to the incremental learning process, but it is due to the  criteria we need to trigger this update process successfully. Given an OCSVM model constructed from the healthy training data, the calculated decision value using Equation \ref{dv2} will decide whether a new event is healthy or not. When this decision value is positive i.e., healthy, then the KKT conditions remain satisfied when this new datum is added to the training data. Thus no model update is required. On the  other hand, when this decision value is negative, we need to know whether this event is related to damage data or  it is only due to environmental changes such as temperature. If this event is  real damage then we report it without any model update. Nevertheless,  if it is due to environmental changes, we need to add this datum to the training data and update the model coefficients accordingly since this negative decision datum will violate the KKT conditions.  The challenge  now  is how to separate  the environmental changes from real damage.

In this paper, we propose a new criterion to trigger the update process of the online-OCSVM based on the information derived from the location component matrix, we obtain when we decompose the three-way tensor data $\mathcal{X}$. This matrix stores meaningful information about the behavior for each sensor location on the bridge. Intuitively, environmental changes such as temperature will affect all the instrumented sensors on the bridge similarly. However, real damage would affect a particular sensor location and the ones close by. To implement this approach, we need to find an efficient solution for online-CP decomposition which will be discussed in the following section.

\subsection{Nesterov SGD (NeSGD) for Online-CP Decomposition}
We employ  stochastic gradient descent (SGD) algorithm to perform CP decomposition in online  manner. SGD  has the capability to deal with big data and online learning models. The key element for optimization problems in SGD is defined as:
\begin{eqnarray}\label{eq:sgd}
w =  \underset{w}{argmin} \;    \mathbb{E}[L(x,w)]
\end{eqnarray}
where $L$ is the loss function needs to be optimized, $x$ is a data point and $w$ is a variable. 

\noindent The SGD method solves the above problem defined in Equation \ref{eq:sgd} by  repeatedly updates $w$ to minimize $L(x,w)$. It starts with some initial value of $w^{(t)}$ and then repeatedly performs the update as follows:
\begin{eqnarray}\label{eq:sgdu}
w^{(t+1)} :=    w^{(t)}   + \eta   \frac{\partial L}{\partial w } (x^{(t)} ,w^{(t)} )
\end{eqnarray}
where $\eta$ is the learning rate and  $\frac{\partial L}{\partial w }$ is the partial derivative of the loss function  with respect to the
parameter we need to minimize i.e. $w$. 

\noindent In the setting of tensor decomposition, we need to calculate the  partial  derivative of the loss function $L$ defined in Equation \ref{eq:als} with respect to the three modes $A, B$ and $C$ alternatively as follows:
\begin{eqnarray}\label{eq:partial}
\frac{\partial L}{\partial A }(X^{(1)}; A)  =   (X^{(1)} -   A \times  (C \circ B)) \times (C \circ B) \nonumber\\
\frac{\partial L}{\partial B }(X^{(2)}; B)  =   (X^{(2)} -   B  \times (C \circ A)) \times (C \circ A)\\
\frac{\partial L}{\partial C }(X^{(3)}; C)  =   (X^{(3)} -   C \times  (B \circ A)) \times (B \circ A)\nonumber
\end{eqnarray}
where $X^{(i)}$ is an unfolding matrix of tensor $\mathcal{X}$ in mode $i$. The gradient update step for $A, B$ and $C$ are as follows:

\begin{eqnarray}\label{eq:update}
A^{(t+1)} :=    A^{(t)}   + \eta^{(t)}   \frac{\partial L}{\partial A } (X^{(1, t)} ;A^{(t)} )  \nonumber\\
B^{(t+1)} :=    B^{(t)}   + \eta^{(t)}   \frac{\partial L}{\partial B } (X^{(2, t)} ;B^{(t)} )  \\
C^{(t+1)} :=    C^{(t)}   + \eta^{(t)}   \frac{\partial L}{\partial C } (X^{(3, t)} ;C^{(t)} )  \nonumber
\end{eqnarray}

The rational idea of GSD algorithm depends only on the gradient information of  $\frac{\partial L}{\partial w }$. In such a non-convex setting, this  partial derivative may encounter data points with  $\frac{\partial L}{\partial w } = 0$ even though it is not at a global minimum. These data points are known as saddle points  which may detente the optimization process to reach the desired local  minimum if not escaped  \cite{ge2015escaping}. These saddle points  can be  identified by studying the second-order derivative  (aka Hessian)  $\frac{\partial L}{\partial w }^2$. Theoretically, when the $\frac{\partial L}{\partial w }^2(x;w)\succ  0$, $x$ must be a local minimum; if $\frac{\partial L}{\partial w }^2(x;w) \prec 0$, then we are at a local maximum; if $\frac{\partial L}{\partial w }^2(x;w)$ has both positive and negative eigenvalues, the point is a saddle point. The second order-methods guarantee  convergence, but the  computing of Hessian matrix $H^{(t)}$ is  high,  which makes the method infeasible for high dimensional data and online learning. Ge \etal \cite{ge2015escaping} show that saddle points are very unstable and can be escaped if we slightly perturb them with some noise. Based on this, we use the perturbation approach which adds Gaussian noise to the gradient. This reinforces the next update step to start moving away from that saddle point toward the correct direction. After a random perturbation, it is highly unlikely that the point remains in the same band and hence it can be efficiently escaped (i.e., no longer a saddle point)\cite{jin2017escape}. Since we are interested in the fast optimization process due to online settings, we further incorporate Nesterov's method into the PSGD algorithm to accelerate the convergence rate. Recently, Nesterov's Accelerated Gradient (NAG) \cite{nesterov2013introductory} has received much attention for solving convex optimization problems \cite{guan2012nenmf,nitanda2014stochastic,ghadimi2016accelerated}. It introduces a smart variation of momentum that works slightly better than standard momentum. This technique modifies the traditional SGD by introducing velocity $\nu$ and friction $\gamma$, which tries to control the velocity and prevents overshooting the valley while allowing faster descent. Our idea behind Nesterov's is to calculate the gradient at a next position that we know our momentum will reach it instead of calculating the gradient at the current position. In practice, it performs a simple step of gradient descent to go from $w^{(t)} $ to $w^{(t+1)}$, and then it shifts slightly further than $w^{(t+1)}$ in the direction given by $\nu^{(t-1)}$. In this setting, we model the gradient update step with NAG as follows: 
\begin{eqnarray}\label{eq:nagNe}
A^{(t+1)} :=    A^{(t)}   + \eta^{(t)}   \nu^{(A, t)} +  \epsilon - \beta ||A||_{L_1} 
\end{eqnarray}
where
\begin{eqnarray}\label{eq:velNe}
\nu^{(A, t)} :=    \gamma \nu^{(A, t-1)}  + (1-\gamma)  \frac{\partial L}{\partial A } (X^{(1, t)} ,A^{(t)} ) 
\end{eqnarray}
where $\epsilon$ is a Gaussian noise, $\eta^{(t)}$ is the step size,  and $||A||_{L_1}$ are the regularization and penalization parameter into the $L_1$ norms to achieve smooth representations of the outcome and thus bypassing the perturbation surrounding the local minimum problem. The updates for $(B^{(t+1)} , \nu^{(B, t)})$ and $(C^{(t+1)}  ,\nu^{(C, t)} )$ are similar to the aforementioned ones.
With NAG, our method achieves a global convergence rate of $O(\frac{1}{T^2})$ comparing to $O(\frac{1}{T})$  for traditional gradient descent. Based on the above models, we present our NeSGD algorithm \ref{NeCPD}.
\begin{algorithm}
	\caption{  NeSGD algorithm}
	\label{NeCPD}
	\textbf{Input}: Tensor $X \in \Re^{I \times J \times K}  $ , number of components $R$\\
	\textbf{Output}: Matrices  $A \in \Re^{I \times R}$, $B \in \Re^{J \times R} $ and  $ C \in \Re^{K \times R}$
	\begin{enumerate}
		\item[1:] Initialize $A,B,C$
		\item[2:] Repeat
		{\setlength\itemindent{6pt}
			\item[3:] Compute the partial derivative of $A, B$ and $C$ using Equation \ref{eq:partial}
			\item[3:] Compute $\nu$ of $A, B$ and $C$ using Equation \ref{eq:velNe}
			\item[4:] Update the matrices  $A, B$ and $C$ using Equation \ref{eq:nagNe}	
		}
		\item[6:] until convergence
	\end{enumerate}
	
\end{algorithm}

\subsection{Tensor-based Advised Decision Values}
In this paper, we introduce a new criterion to trigger the update process  when the online OCSVM model generates a negative decision value for a new sample $c^{t+1}$.  Based on the information derived from the location component matrix $B^{(t+1)}$ which we obtain when we decompose a three-way tensor data $\mathcal{X}^{t+1}$, we generate an advised decision score for a new negative datum based on the average distance from a sensing location matrix (a row in (\textbf{\textit{B}})) to the k nearest neighbouring (knn) locations. A big change in this score of a sensing
location indicates a change in sensor behaviour which might be due to occurred damage or environmental changes.  If all the knn scores behave differently, then this indicates that the negative decision value is due to environmental changes. Therefore, we add this new datum to the training data and update the model coefficients accordingly. Otherwise, we report  $c_i^{t+1}$ as an anomaly data point.
This algorithm is described as follows: given the location matrix  $B^{t+1}$, the unit vector of each point $b_j (j = 1, \ldots, n)$ with its $k$ closest points $b_k$ is computed as follows: 
\begin{eqnarray}\label{norm}
v_{j}^{k} = \dfrac{b_j-b_k}{ \lVert b_j-b_k \rVert}
\end{eqnarray}
Then we estimate the change occurred in sensors behaviors based on the absolute difference   between $B^{t+1}$  and $B^{t}$ using the following equation
\begin{eqnarray}\label{rate}
\mathcal{P}_i(C^{t+1}) =     \frac{1}{n}\sum_{n=1}^{j}\lvert b_j^{t+1}-b_j^{t} \rvert >  \gamma
\end{eqnarray}
where $\mathcal{P}_i$ represents the probability of that negative decision value of $c_i^{t+1}$  is being related   to environmental  changes, and $\gamma$ is a small value represents acceptable change.  In this work, we set-up a 90\% confidence interval for $\mathcal{P}_i(C^{t+1})$  to judge whether a new datum $c_i^{t+1}$ belongs to the healthy sample or not. In fact, this confidence interval is based on the physical layout of the sensor instrumentation. Algorithm \ref{TDV} illustrates the process of generating the tensor-based advised decision values:

\begin{algorithm}[!h]
	\caption{ Tensor-based advised decision values method.}
	\label{TDV}
	\textbf{Input}: A set of $n$ sensors $B^{(t+1)}=\{{b_j}\}_{i=1}^n$\\
	For each sample ${b_j}$  in $B^{(t+1)}$
	\begin{enumerate}
		\item[(a)]Find the $k$ closest points to $b_j$: $ j= 1, \dots,k$.
		\item[(b)] Calculate the unit vectors $v_j^k$ of  $b_{j}$ according to (\ref{norm}).
		\item[(c)] Calculate $\mathcal{P}_i$ according to (\ref{rate}).	
		
		\item[(d)] Adjust the decision value as follows:		
		
		$g(c^{t+1})$ = \Bigg\{\begin{tabular}{lp{2em}l}
			$\vert g(c^{t+1}) \vert$ & &   $ \mathcal{P}_i(c^{t+1}) \geq 0.9$\\
			$ g(c^{t+1})  $ & &  $\mathcal{P}_i(c^{t+1}) < 0.9$
		\end{tabular}
		
		\item[(e)] Update the model coefficients
	\end{enumerate}
\end{algorithm}

\subsection{Model Coefficients Update}
The model solution of OCSVM is basically composed of two coefficients denoted by $\alpha_i$ and $\rho$.  When a new datum $c_i^{t+1}$  arrives with an advised decision value $a(x_c)>0$, the model coefficients must be updated to get the new solution $\alpha_{i+c}$ and $\rho*$, which should  preserve  the KKT conditions (see Equation \ref{kkt}). We initially assign a zero value to $\alpha_c$ and then  start looking for the largest possible increment of $\alpha_c$ until  its $g(x_c)$
becomes zero, in this case $x_c$ is added to the support vectors set $S$. If $\alpha_c$  equals to $1$, $x_c$ is added to the set of the error vectors  $E$.

The difference between the two states of OCSVM is shown in the following
equation:

\begin{eqnarray}\label{pd}
\Delta g_i = \phi(x_i,x_c) \Delta \alpha_c + \sum_{j \in S} \phi(x_i, x_j) \Delta \alpha_j + \Delta \rho.
\end{eqnarray}

Since $g_i = 0$ $ \forall i \in S$, we can write Equation \ref{pd} in a matrix form as follows:

\begin{eqnarray}\label{pdm}
\underbrace{ \Bigg[
	\begin{tabular}{cc}
	0 &    $1^T$ \\
	1 &   $\phi_{s,s}$
	\end{tabular}
	\Bigg]}_Q
\underbrace{	 \Bigg [
	\begin{tabular}{c}
	$\Delta \rho$\\
	$\Delta \alpha_s$
	\end{tabular}
	\Bigg ] }_{\Delta \tilde{\alpha_s}}
=-
\underbrace{ \Bigg [
	\begin{tabular}{c}
	1\\
	$\phi_{s,c}$
	\end{tabular}
	\Bigg ] }_{\eta_c} \Delta \alpha_c
\end{eqnarray}
\vspace{-6mm}
\begin{eqnarray}\label{le}
\Delta \alpha_s = \underbrace{- Q^{-1} \eta_c}_\beta  \Delta \alpha_c
\end{eqnarray}

By substituting $\alpha_s$ in the  partial derivate equation \ref{pd}, we obtain:
\vspace{-2mm}
\begin{eqnarray}\label{pdb}
\Delta g_i = \phi(x_i,x_c) \Delta \alpha_c + \sum_{j \in S} \phi(x_i, x_j) \beta_j\Delta \alpha_c + \beta_0.
\end{eqnarray}
\vspace{-5mm}
\begin{eqnarray*}
	\Delta g_i  = \gamma_i  \Delta \alpha_c
\end{eqnarray*}
\vspace{-3mm}
where
\vspace{-3mm}
\begin{eqnarray*}\label{gamma}
	\gamma_i = \phi(x_i,x_c)  + \sum_{j \in S} \phi(x_i, x_j) \beta_j + \beta_0.
\end{eqnarray*}

The goal now is to determine the index of the sample $i$ that leads to the minimum
$\Delta \alpha_c$. As in \cite{cauwenberghs2000incremental}, five cases must be
considered to manage the migration of the sample between the three sets $S$, $E$ and
$R$, and calculate $\Delta \alpha_c$.

\begin{enumerate}
	
	\item $\Delta \alpha_c^{1} = \min_i \frac{1-\alpha_i}{\beta_i}$, $\forall i \in S$ and $\beta_i > 0 $.  \\
	$   \hspace*{1cm} \Delta \alpha_c^{1}$ leads to the minimum $ \Delta \alpha_c $
	$\rightarrow$ Move $i$  from $S$ to $E$.
	
	\item  $\Delta \alpha_c^{2} = \min_i \frac{-\alpha_i}{\beta_i}$, $\forall i \in S$ and  $\beta_i < 0 $\\
	$\hspace*{1cm} \Delta \alpha_c^{2}$ leads to the minimum $ \Delta \alpha_c $
	$\rightarrow$ Move $i$  from $S$ to $R$.
	
	\item   $\Delta \alpha_c^3$ = $\min_i \frac{-g_i}{\gamma_i}$, 
	$ \forall i \in E$ and  $\gamma_i > 0$ or  $\forall i \in R$ and  $\gamma_i < 0$.\\ 
	$\hspace*{1cm} \Delta \alpha_c^{3}$ leads to the minimum $ \Delta \alpha_c $
	$\rightarrow$   Move $i$  from $E$ or $R$ to $S$. 	  
	
	\item $\Delta \alpha_c^4 = \frac{-g_c}{\gamma_c}$, $i$ is the index of $x_c$. \\
	$\hspace*{1cm}\Delta \alpha_c^{4}$ leads to the minimum $ \Delta \alpha_c $	    
	$\rightarrow$   Move $x_c$  to $S$, terminate.
	
	\item  $\Delta \alpha_c^5 = 1-\alpha_c$, $i$ is the index of $x_c$.\\
	$\hspace*{1cm}\Delta \alpha_c^{5}$ leads to the minimum $ \Delta \alpha_c $	  
	$\rightarrow$   Move $x_c$  to $E$, terminate.
	
\end{enumerate}

The next step after the migration is to update the inverse matrix $Q^{-1}$. Two
cases to consider during the update: extending $Q^{-1}$ when the determined
index $i$  joins $S$, or compressing when index $i$  leaves $S$. Similar to
\cite{laskov2006incremental}, we applied the Sherman-Morrison-Woodbury formula to
obtain the new matrix $\tilde{Q}$. We repeat this procedure until the \ index $i$
is related  to the new example $x_c$.

\section{Experimental Results}
\label{sec:Experiement}
We conduct all our experiments  using an 	Intel(R) Core(TM) i7 CPU 3.60GHz with 16GB memory. We use R software to implement our algorithms  with the help of the two packages; the  \textbf{rTensor}
and the \textbf{e1071} for tensor tools and one-class model.

\subsection{Experiments on Synthetic Data}

\subsubsection{NeSGD convergence}
\begin{figure}[!t]
	\centering
	\includegraphics[scale=0.6]{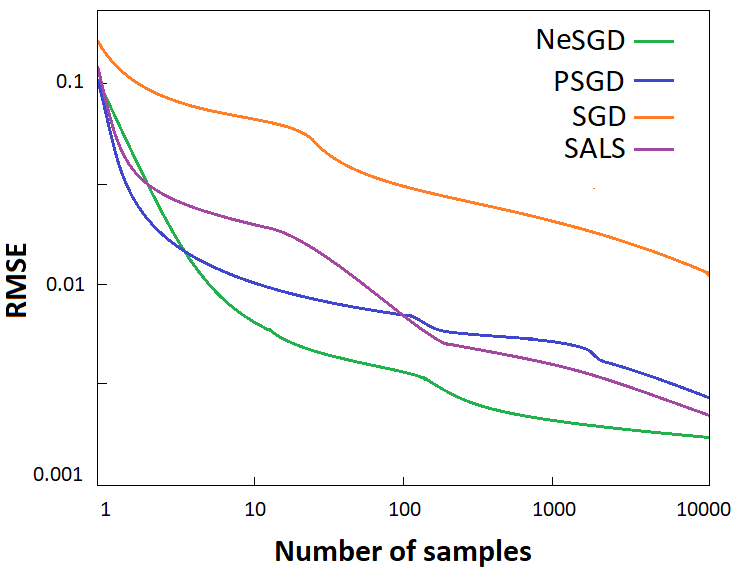}
	\caption{ Comparison of algorithms in terms of robustness and convergence. }
	\label{conv}
\end{figure}

Our initial experiment was to ensure the convergence of our NeSGD algorithm and compare it to other state-of-the-art methods such as SGD, PSGD,  and SALS algorithms in terms of robustness and convergence. We generated a synthetic long time dimension   tensor $ \mathcal{X} \in \Re^{60 \times 12 \times 10000}$ from 12 random   loading matrices ${M}_{i=1}^{12} \in   \Re^{10000 \times 60}$ in which entries were drawn from uniform distribution $\mathcal{D} [0,1]$. We evaluated the performance of each method by plotting the number of samples $t$ versus the root mean square error (RMSE). For all experiments we use the learning rate $\eta^{(t)} =  \frac{1}{1 + t}$. It  can be clearly seen from Figure \ref{conv}  that  NeSGD outperformed the SGD and PSGD algorithms in terms of convergence and robustness.  Both SALS and NeCPD converged to a small RMSE but it was lower and faster in NeSGD.

\subsubsection{Tensor-based analysis}
\begin{figure}[!t]
	\centering
	\captionsetup[subfloat]{}
	\subfloat[The resultant decision boundary of the  OCSVM trained on the 500 healthy samples.]{{\includegraphics[height=1.6in,width=1.8in]{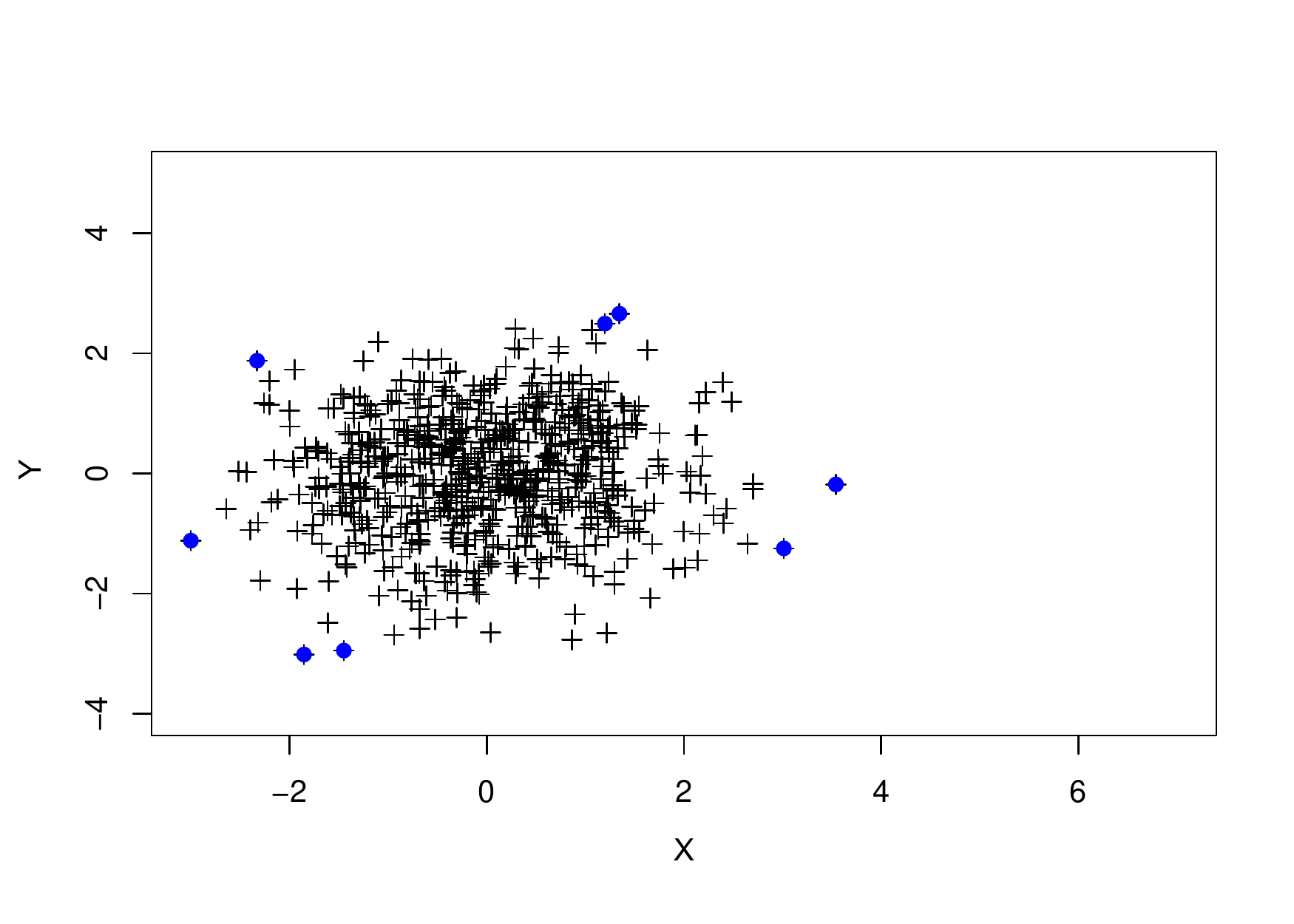} }}%
	\qquad
	\subfloat[An example result of the obtained KNN score generated from the $B^{(t)}=\{{b_j}\}_{i=1}^{12}$ ]{{\includegraphics[height=1.6in,width=2.6in]{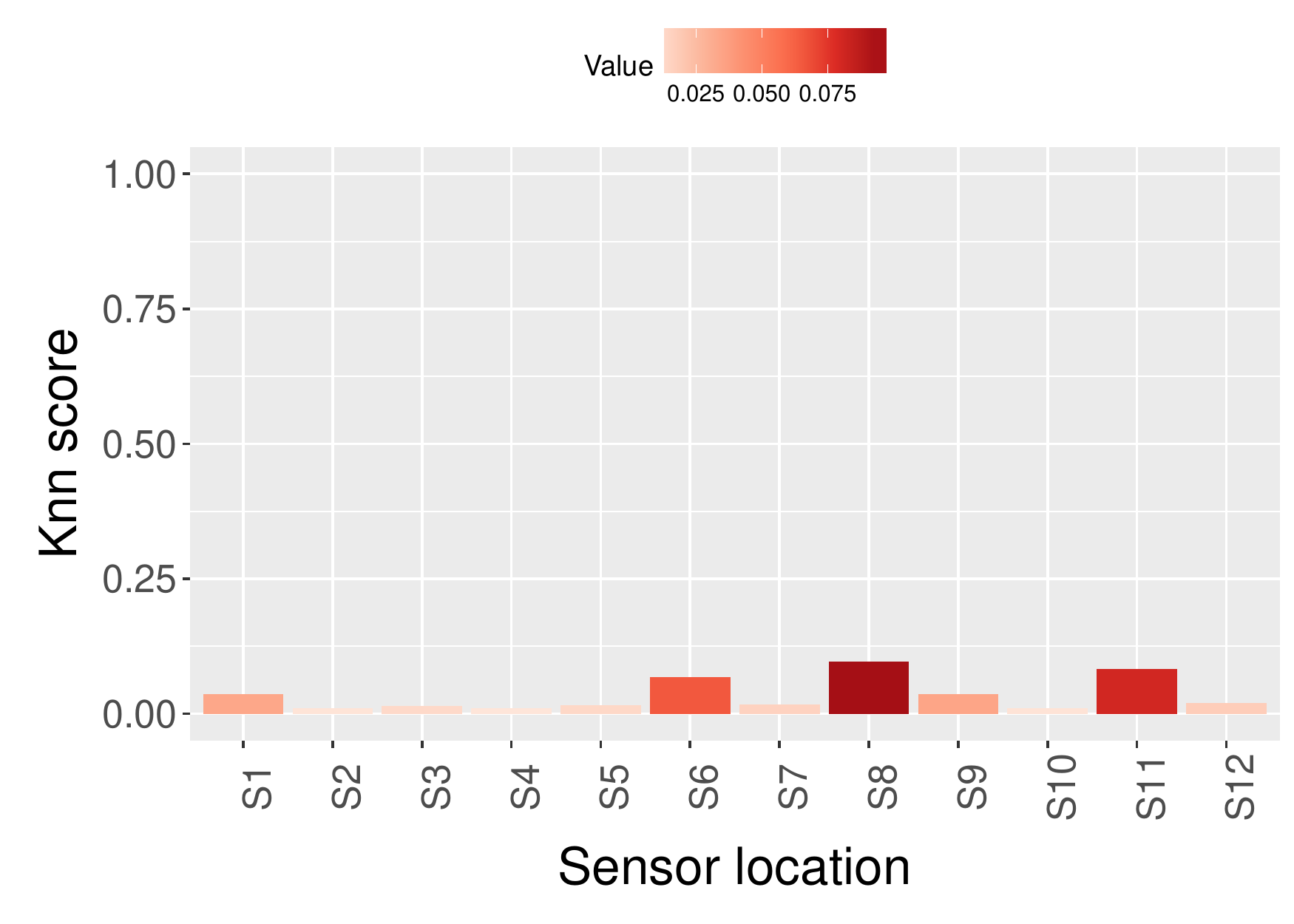} }}%
	\caption{ Experimental results using the Synthetic data.}%
	\label{offline_syn}%
	
\end{figure}

The same synthetic data  is used here but with some modifications to evaluate the performance of the proposed  method of tensor-based advice decision value. We initially emulate environmental  changes on the last 300 samples by modifying the random generator setting $[\mu,\sigma ]$, where $\mu$ is the mean and $\sigma$ is the standard deviation,  in all $M's$  since   environmental  changes which will naturally affect all sensors,   where each matrix in ${M}_{i=1}^{12} \in   \Re^{1000 \times 60}$ represents one source (i.e location). The first 700 samples were generated under the same environmental conditions in which 500 samples are used to form the training tensor  $ \mathcal{X} \in \Re^{60 \times 12 \times 500}$. The NeSGD is initially  applied to decompose the tensor  $ \mathcal{X}$ into three matrices $ A, B,$ and $C$ given $R=2$, and the $C$ matrix is then used to learn the OCSVM. The remaining 200 samples where fed sequentially to the online NeSGD and OCSVM in addition to the environmental affected 300 samples. For each incoming datum, we compute $A^{t+1}, B^{t+1}$ and $C^{t+1}$ which  is presented to the online OCSVM algorithm to calculate its health score. If that datum is predicted as healthy then the model is not updated. However, if that datum is predicted as unhealthy, then we compute the tensor-based decision value $\mathcal{P}_i(C^{t+1})$ using Equation \ref{rate}. If its advice decision value is greater than $\gamma$ then we incorporated this new datum into the training data and we updated the model's coefficients accordingly. Figure \ref{offline_syn} shows the initial 500 training samples where the blue dots represent the resultant support vectors of the offline model. Figure \ref{porto} shows the resultant decision boundary after we incrementally updated the OCSVM model. As it can be seen, all the healthy samples were successfully incorporated into the  model since these samples were slightly different from the training samples due to the environmental changes. We can also observe that the KNN score for all sensor were significantly increased which indicates the negative decision values were due to environmental changes. Further, it shows how the decision boundary  grew over time and  incorporated  new healthy samples. 

\begin{figure}[!t]
	\centering
	\captionsetup[subfloat]{}
	\subfloat[The resultant decision boundary of the  incremntally updated OCSVM after processing the 500 test samples.]{{\includegraphics[height=1.6in,width=1.8in]{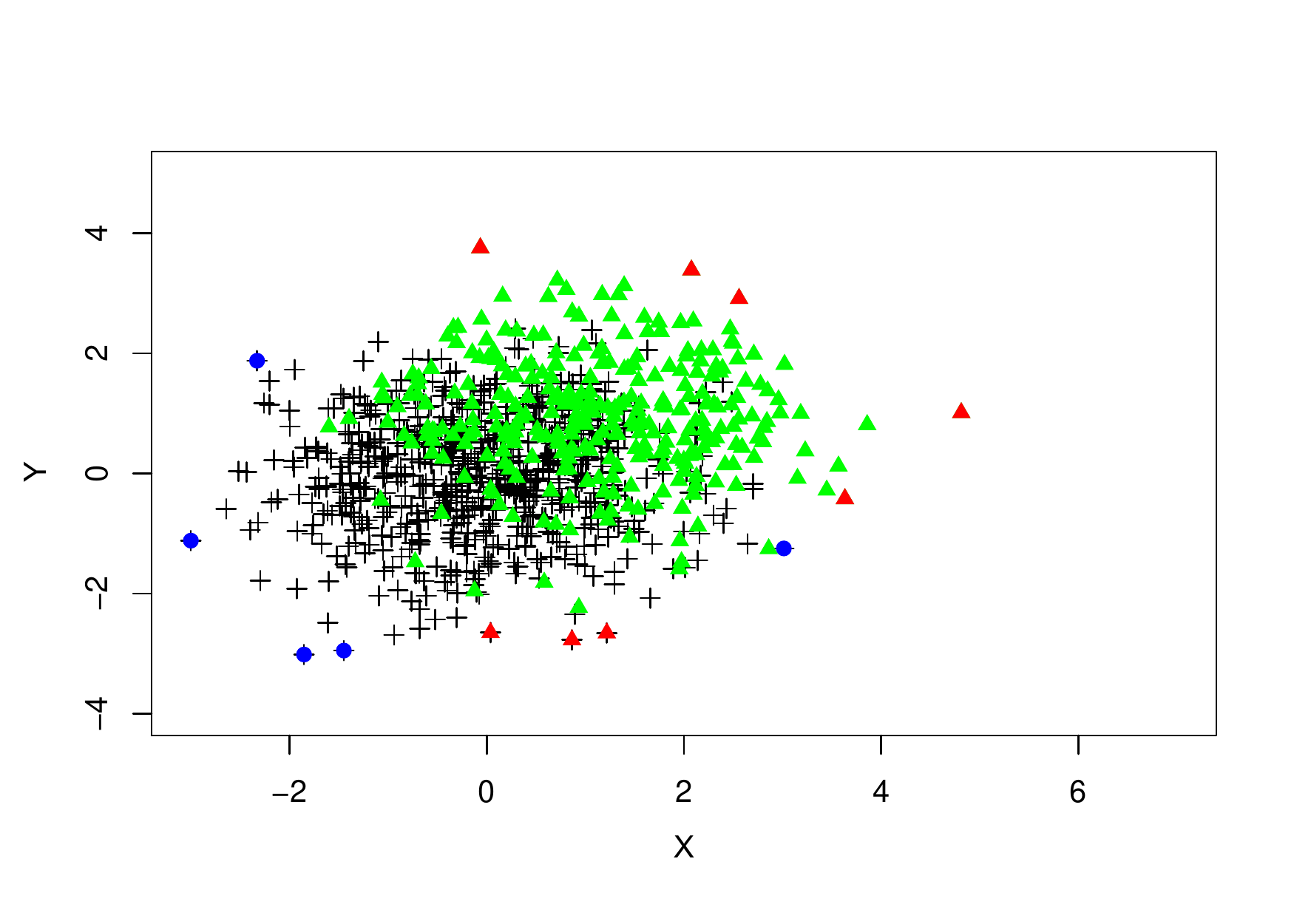} }}%
	\qquad
	\subfloat[An example result of the obtained KNN score generated from the $B^{(t+201)}=\{{b_j}\}_{i=1}^{12}$ ]{{\includegraphics[height=1.6in,width=2.6in]{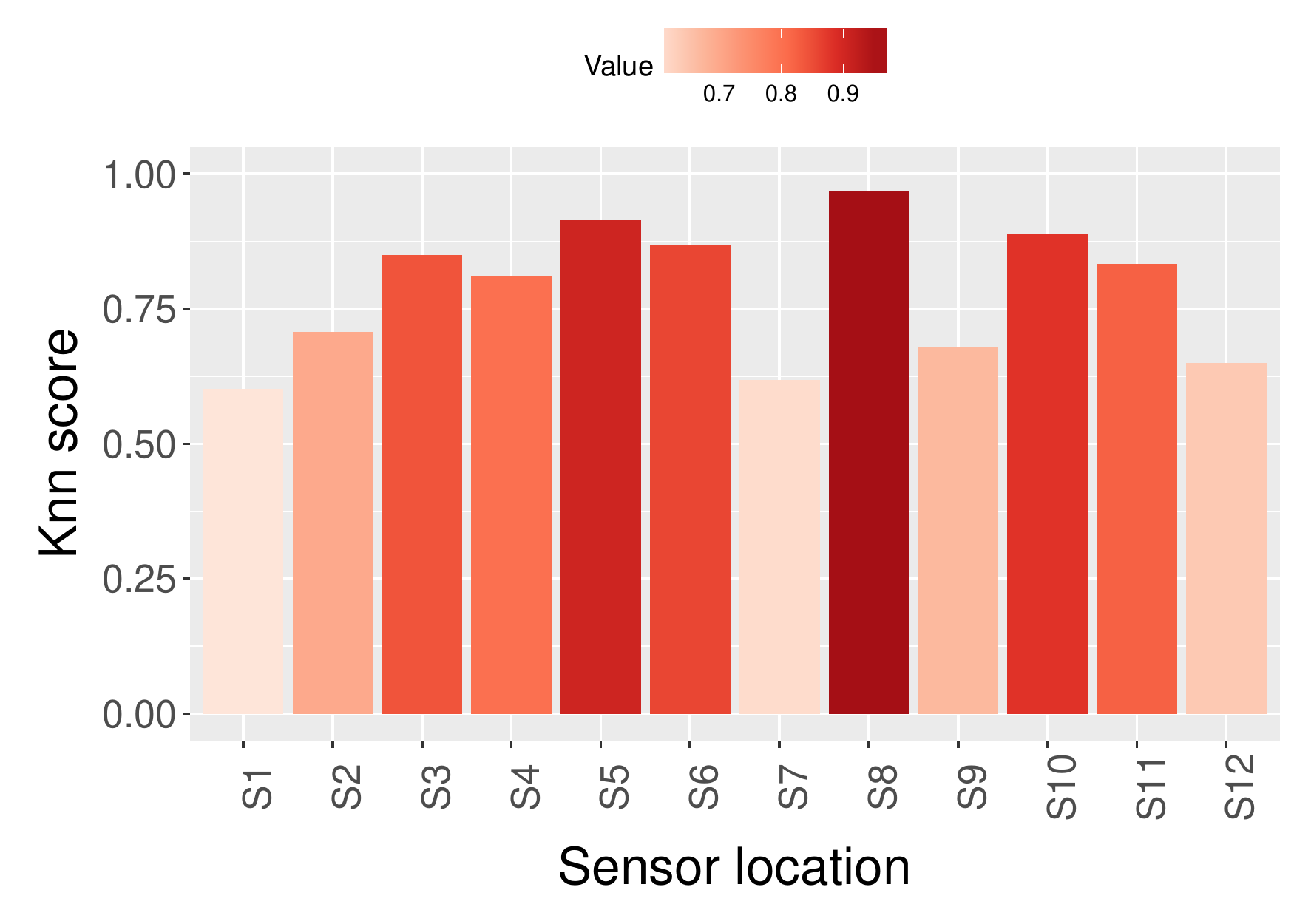} }}%
	\caption{ Experimental results using the Synthetic data.}%
	\label{porto}%
	
\end{figure}

\subsection{Case Studies in Structural Health Monitoring}

To further validate our model, we evaluate the performance of our Tensor-based advice decision values for incremental OCSVM when it is applied to SHM. The evaluation is based on real SHM data collected from two case studies namely (a) the Infante D. Henrique Bridge in Portugal and (b) AusBridge a major Bridge in Australia (the actual Bridge name cannot be published due to a data confidentiality agreement).


\subsubsection {Infante D. Henrique Bridge Data}
The SHM data is collected continuously from the Infante D. Henrique Bridge bridge (shown in Figure \ref{porto} (a)) over a period of two years (from September 2007 to September 2009) \cite{comanducci2016vibration,magalhaes2012vibration}. The data collection was carried out continuously by instrumenting the bridge with 12 force-balance highly sensitive accelerometers to measure acceleration. Every 30 minutes, the collected acceleration measurements were retrieved from these sensors and input into an operational modal analysis process to determine the natural frequencies (the model parameters) of the bridge. This process resulted in 120 natural frequencies which used as the characteristic features in our study. Thus, this dataset consists of $ 2 \times 365 \times 48 $ (35,040) samples each with 24 features. The resultant matrices from the 12 sensors   were fused in a tensor $X \in \Re^{35,040 \times 120 \times 12}$.  Besides acceleration measurements, temperatures were also recorded every 30 minutes because natural frequencies of the bridge can be influenced by environmental conditions \cite{comanducci2016vibration,magalhaes2012vibration}.

Using this dataset, we run two-phase experimental analysis; In the first phase, we used the first two months of this data (i.e., 2,230 samples September-October 2007) to fuse them in a tensor  $X \in \Re^{2,230 \times 120 \times 12}$ which was decomposed using NeSGD method (see Algorithm \ref{NeCPD}) into three matrices  $A  \in \Re^{12 \times 2}$,  $B \in \Re^{120 \times 2}$, and  $C \in \Re^{2,230 \times 2}$.  The $C$ matrix is then used to construct an offline OCSVM model. We used the remaining 35,040 data samples to evaluate the   offline OCSVM model without applying the tensor-based advice decision value method. This analysis resulted in a very high false alarm rate of 44.8\%. This result demonstrates the significant effect of environmental factors, particularly the temperature, on the natural frequencies of the data collected from the bridge. 
In the second phase of this experiment, we run our proposed tensor-based advised incremental OCSVM algorithm on the same test dataset. Here, we calculated the health score for each incoming sample using equation \ref{dv2}. The algorithm then continues if the sample has been correctly classified, otherwise the tensor-based advice decision value has to be calculated to determine whether the model coefficients will have to be updated or not based on Equation \ref{rate}.

The results of these experiments are shown in figure \ref{porto} (b); mainly resulted in a false alarm rate per month for the (tensor-based,  threshold-based, self-advised) online OCSVM and the offline OCSVM. This figure also shows the monthly average temperature for the period of September 2007 and October 2007, which was used for constructing the offline OCSVM, for demonstrating the environmental conditions of the constructed model. As depicted in this figure, although the false alarm rate of our tensor-based online incremental model was high at the start of the experiment, it had decreased gradually until it reached close to zero as new arriving data augmented into the model. 

By experimenting with the dataset during the period June 2008 - December 2008, our tensor-based online model recorded an above 10\% false alarm rate. This period  time belongs to the extreme temperature  conditions which have not been previously experienced at that time  point in the dataset. The self-advised method produces comparable
results to tensor-based ones but with lower accuracy. In contrast to this, the threshold-based online OCSVM and offline OCSVM showed continuous some fluctuation in the false alarm rates in correlation with the monthly record temperature. Specifically, our tensor-based  online model generated a very low false alarm rate (close to 0\%) during the months which had temperature values that are significantly different from the temperature values recorded during the training period (i.e., September - October 2007 vs. September - October 2008). On the other hand, very high false alarm rates (close to 100\%) were generated by the offline OCSVM model during the same time period.

From this case study, we conclude that environmental changes, which are captured using natural frequencies feature, can significantly influence OCSVM models. Our experiments with the real Infante D. Henrique Bridge dataset have demonstrated that our tensor-based  online model is able to catch such environmental changes in the features.  In this regard, the proposed method makes more accurate updated to the learning models compared to the (self-advised and threshold-based) online and offline models.

\begin{figure}[!t]
	\centering
	\captionsetup[subfloat]{}
	\subfloat[Infante D. Henrique Bridge \cite{magalhaes2012vibration}.]{{\includegraphics[height=1.6in,width=1.8in]{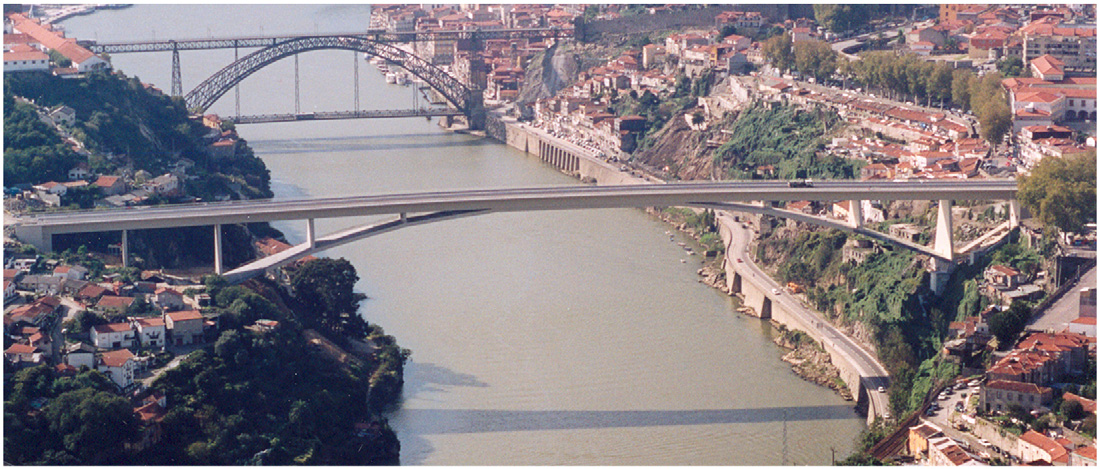} }}%
	\qquad
	\subfloat[Error rates of the online models versus offline models in relation to the changes in the temperature.]{{\includegraphics[height=1.6in,width=3.2in]{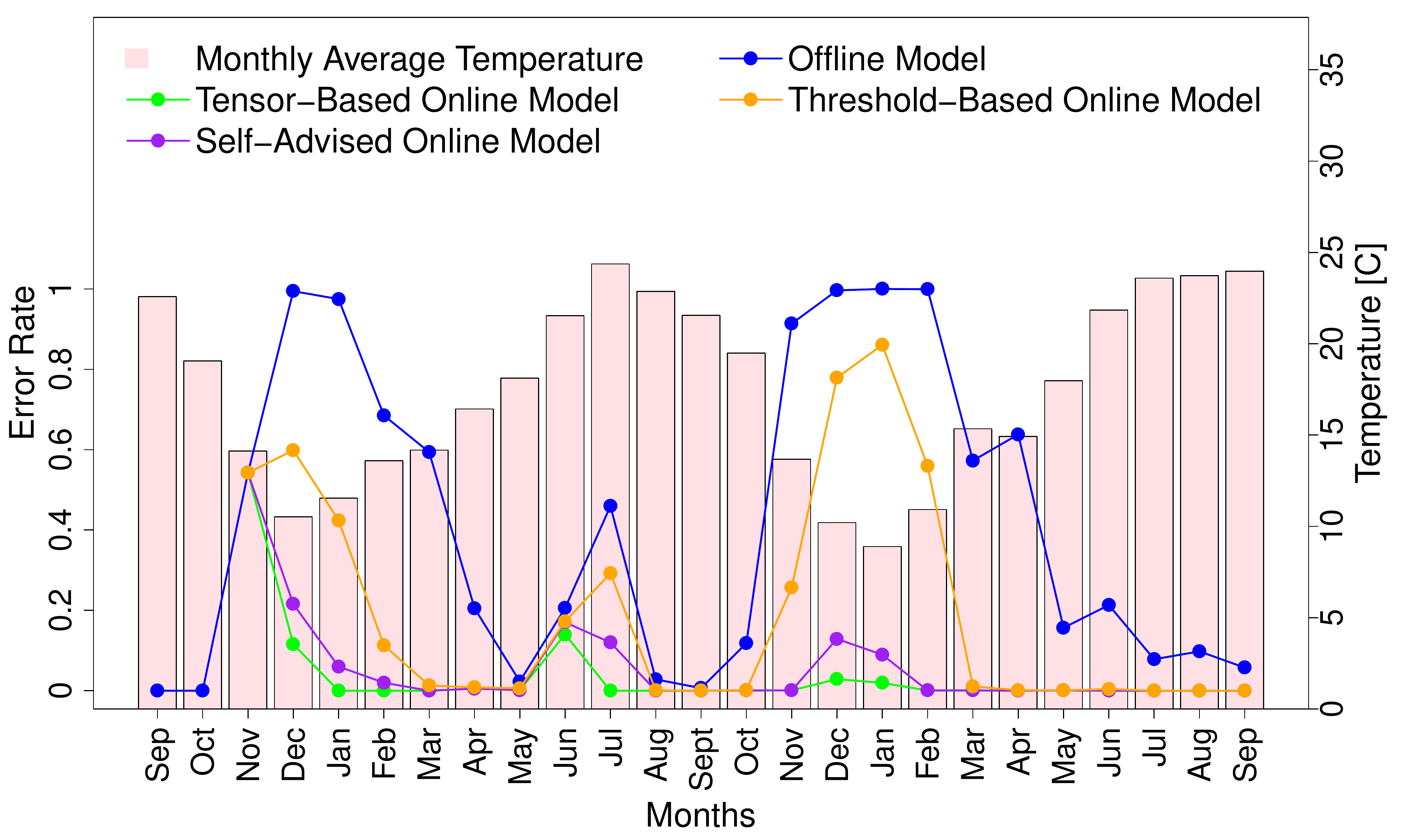} }}%
	\caption{ Experimental results using data from the Infante D. Henrique Bridge.}%
	\label{porto}%
\end{figure}

\subsubsection{The AusBridge}
In this case study, we collect acceleration data from the AusBridge using tri-axial accelerometers connected to a small computer under each jack arch.  Every vehicle passed over a given jack arch it triggers an event on it. The sensors attached to that jack arch collected all the resulted vibrations for a duration of 1.6 seconds with a sampling rate of 375 Hz. As a result, each sensor captures 600 samples per event. The collected samples were further normalized and transformed into 300 features in the frequency domain.  

\begin{figure}[!t]
	\centering
	\captionsetup[subfloat]{}
	\subfloat[Comparison of average error rates on the eleven nodes for the batch and online OCSVM.]{{\includegraphics[height=1.5in,width=2.2in]{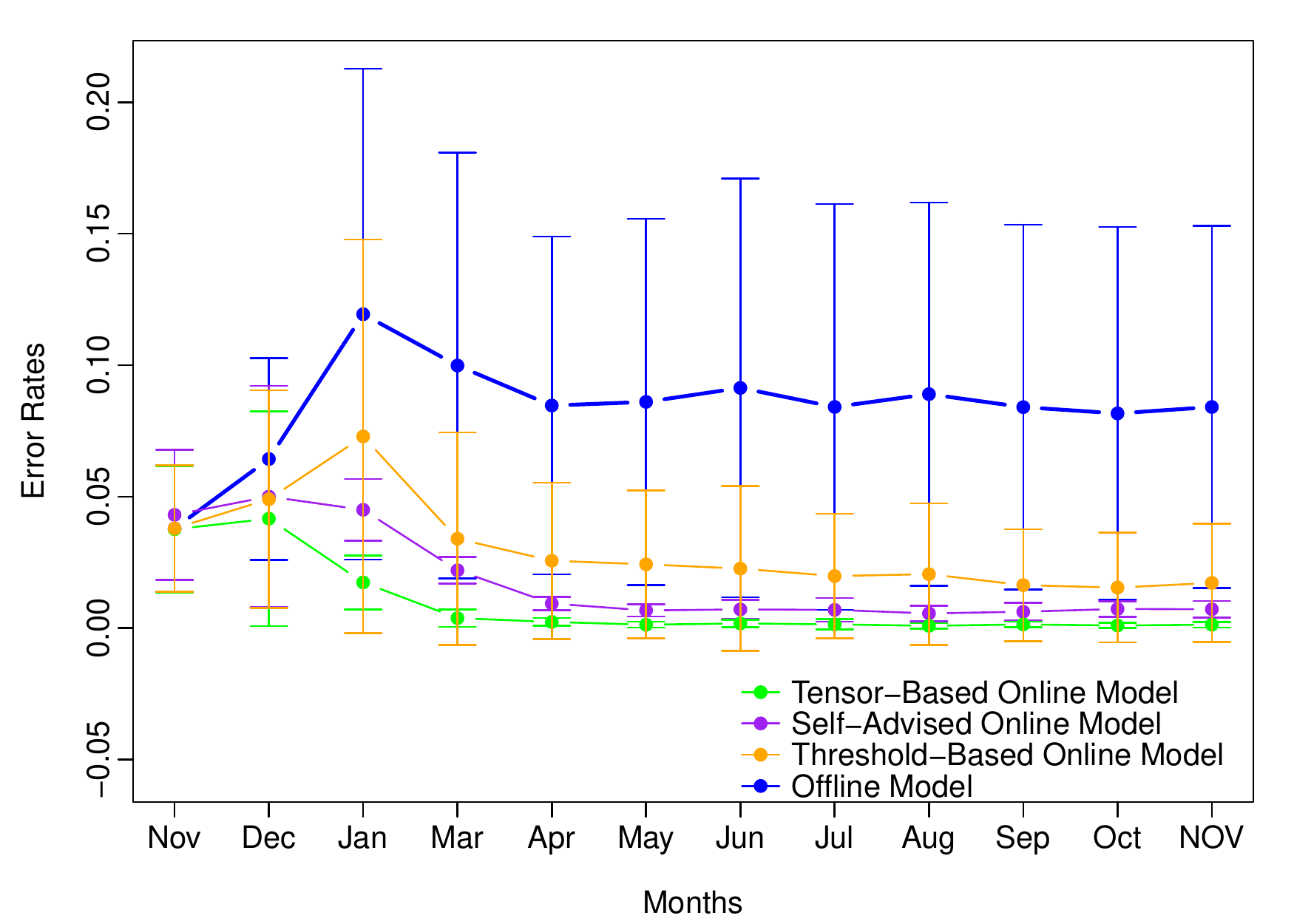} }}%
	\qquad
	\subfloat[Comparison of average detection accuracy between the offline and online OCSVM methods applied to a damaged jack arch.]{{\includegraphics[height=1.7in,width=2.3in]{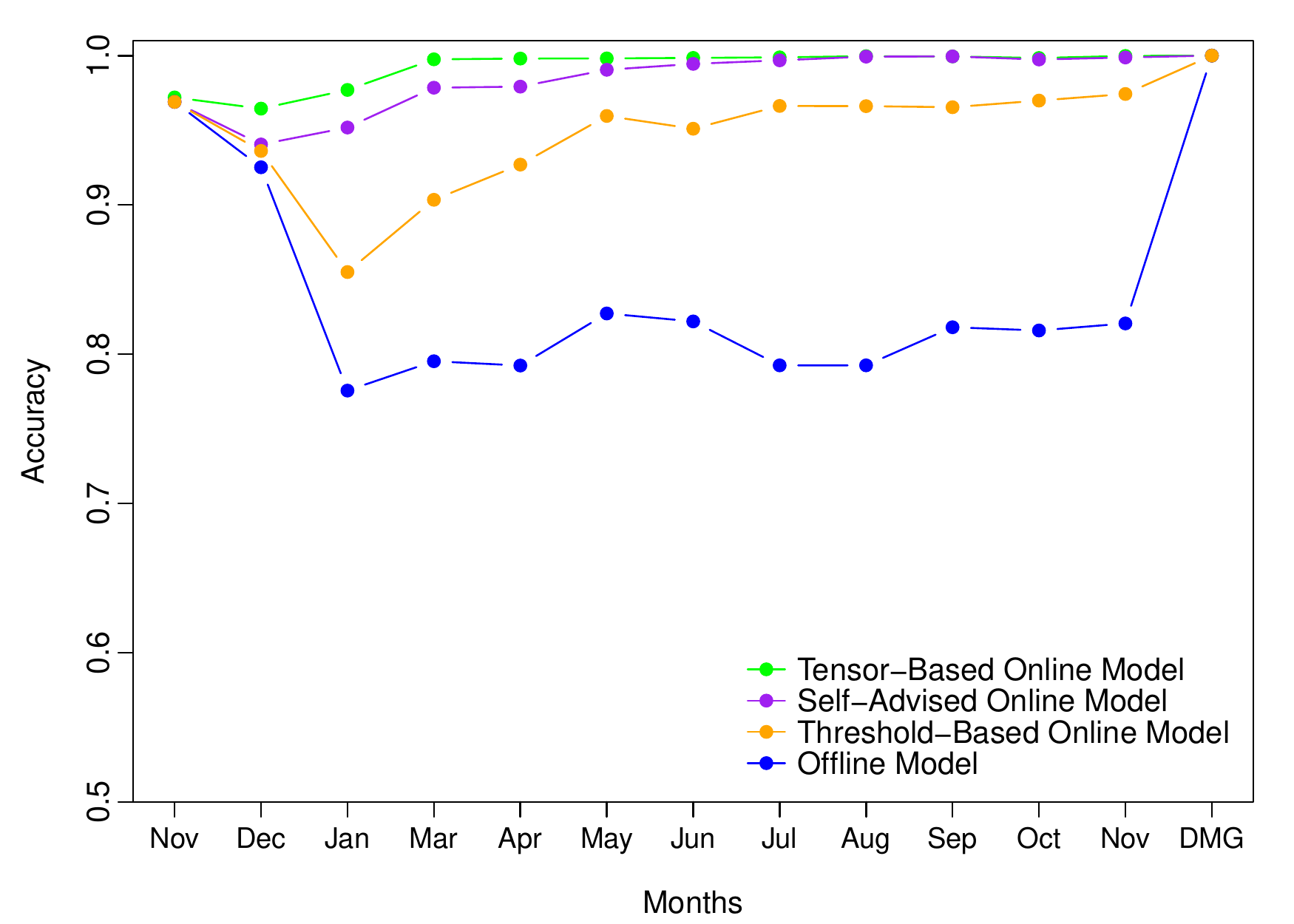} }}%
	\caption{ Experimental results using data from the AusBridge.}%
	\label{nodes}
\end{figure}
We conducted two sets of experiments. The first one used a data of a total of 22,526 samples collected  from 11 jack arches (nodes)  during October 2015 and November 2016\footnote{The data for February 2016 had known instrumentation problem which appeared and was fixed during that period, and thus it was excluded from this experiment.}. The data from the 11 nodes were fused in a tensor $X \in \Re^{22,526 \times 300 \times 11}$. We train the offline OCSVM model using the data collected in October 2015 (i.e $X \in \Re^{1,689 \times 300 \times 11}$ ). The remaining data of $X \in \Re^{20,837 \times 300 \times 11}$ (November 2015 - November 2016) was sequentially fed to the same model. Using the same dataset, we run these experiments with (self-advised and threshold-based) online OCSVM, offline OCSVM ,  and our tensor-based OCSVM model. 

The false alarm error rates resulted from the experiments with the four OCSVM models are illustrated in Figure \ref{nodes}(a). The values are shown as averages over the 11 nodes for each month of the collected data. As shown, the offline OCSVM model generated an average of about 10\% false alarm rate which means it classified many events as structural damage. Furthermore, the false alarm rates fluctuated with high standard deviations during most of the experiment period. This is not accurate and can lead to inappropriate decisions as all of the nodes being evaluated had no damage during the study period. Unlike the offline OCSVM model, our tensor-based online model resulted in a much lower false alarm rate, average about 0.1\% with narrow standard deviations. The poor performance of the offline OCSVM model can be due to the environmental and operational conditions, which were not captured in the initial frequency features in the one-month training data (October 2016). The influence of these variations on the frequency features had been captured by the threshold-based online OCSVM method and enhanced its false alarm rates. As shown in the fugue, our tensor-based  online OCSVM model resulted in lower false alarm rates compared to the other online models. Although setting a fixed threshold for a single OCSVM might be an easy task, but this is not practical. Such bridges comprised of 100's of jack arches  each of which is linked with an OCSVM model. Thus, setting a fixed threshold would require manual tuning of 100's OCSVM models.

The goal of the second experiment set of the AusBridge case study  was to confirm that even after a long-running period (i.e. 1 year) the updated model from our online OCSVM method still had the ability to detect
structural damage. In the second experiment set of the AusBridge case study, we used the data of an identified damaged jack arch (node). In particular,  we used 12 months of data during which the jack arch was health and 3 months of data while it was damaged. Our goal in this experiment is to confirm the reliability of our model in terms of its ability to detect structural damages even after a long time period (1 year). Using the 12 months dataset, we run the offline, self-advised, threshold-based, and our proposed tensor-based  online OCSVM models. Figure \ref{nodes} (b) shows the resulting accuracy of the four models over the 12-months of healthy data and the 3-months damaged data of the jack arch. As shown, the damage events were detected successfully by all the four models. In addition, our tensor-based  model significantly outperformed the offline OCSVM and threshold-based incremental models by achieving lower false alarm rates.

\section{Conclusion}
\label{sec:Conclusion}

We address the problem of learning from data sensed from networked sensors in IoT environments. Such data exists in a correlated multi-way form and  considered as  non-stationary  due to the on-going variation that often arises from environmental changes over a long period  time. Existing learning models such as OCSVM and traditional matrix analysis methods do not capture theses aspects. Thus, accuracy and performance are significantly affected with such  non-stationary nature. We addressed these problems by proposing a new online CP decomposition named NeSGD and a Tensor-based Online-OCSVM method which employs the online learning technique. The essence of our proposed approach is that it triggers incremental updates to the online OCSVM model based on data received from the location component matrix which maintains important information about sensor's behaviour. We achieved this by incorporating a new criterion received regularly from each sensor which is captured by decomposing the tensor using NeSGD. 

We applied our approach to real-life datasets collected from network of sensors attached to bridges to detect damage (anomalies) in its structure that might result from environmental variations over long time periods. The various experimental results showed that our tensor-based online-OCSVM was able to accurately and efficiently differentiate between environmental changes and actual damage behaviours. Specifically, our tensor-based Online OCSVM method significantly outperformed the self-advised and threshold-based OCSVM models as it scored the lowest false alarm rates and carried more accurate updates  to the learning model. 

It would be interesting to investigate other factors (other than temperature) that may influence anomaly detection in structure health monitoring and other areas. One interesting area is to extend and  apply our Tensor-based Online OCSVM model to other related IoT fields such as smart homes.

\begin{acks}	
The authors wish to thank the Roads and Maritime Services (RMS) in New 	South Wales, Australia for provision of the support and testing facilities for this research work. NICTA is funded by the Australian Government through the Department of Communications and the Australian Research Council through the ICT Centre of Excellence Program. CSIRO's Digital Productivity business unit and NICTA have joined forces to create digital powerhouse Data61. The authors also would like to acknowledge Professor Alvaro Cunha and Professor Filipe Magalhaes at Porto University for providing us with continuous real data from Infante D. Henrique Bridge. Their support is highly appreciated.
\end{acks}

\bibliographystyle{ACM-Reference-Format}
\bibliography{incsvm_journal}

\end{document}